\documentclass[11pt]{article}

\usepackage[preprint]{acl}
\usepackage{times}
\usepackage{latexsym}
\usepackage[T1]{fontenc}
\usepackage[utf8]{inputenc}
\usepackage{microtype}
\usepackage{inconsolata}
\usepackage{graphicx}
\usepackage{amsmath,amsfonts}
\usepackage{array}
\usepackage{booktabs}
\usepackage{multirow}
\usepackage{tabularx}

\usepackage{stfloats}

\usepackage{listings}

\begin{document}

\title{Path-Level Structural and Semantic Alignment for Hierarchical Multi-Label Text Classification}


\author{
  \textbf{Xindi Wang\textsuperscript{1}},
  \textbf{Aifen Xiong\textsuperscript{2}},
  \textbf{Baiyang Wang\textsuperscript{2}},
  \textbf{Xuri Ge\textsuperscript{1}},
  \textbf{Robert Mercer\textsuperscript{3}},
  \\
  \textbf{Frank Rudzicz\textsuperscript{4}},
  \textbf{Yifan Peng\textsuperscript{5}},
  \textbf{Zhumin Chen\textsuperscript{1}},
  \textbf{Hongjun Wang\textsuperscript{2}}
  \\
  \\
  \textsuperscript{1}School of Artificial Intelligence, Shandong University, Jinan, China\\
  \textsuperscript{2}School of Information Science and Engineering, Shandong University, Qingdao, China\\
  \textsuperscript{3}Department of Computer Science, Western University, London, Ontario, Canada\\
  \textsuperscript{4}Faculty of Computer Science, Dalhousie University, Halifax, Nova Scotia, Canada\\
  \textsuperscript{5}Department of Population Health Sciences, Weill Cornell Medicine, New York, NY, USA\\
  {\normalfont
  \textbf{Correspondence:}
  \texttt{chenzhumin@sdu.edu.cn};
  \texttt{hjw@sdu.edu.cn}
  }
}

\maketitle

\begin{abstract}
Hierarchical multi-label text classification (HMTC) requires models to capture both structural dependencies among labels and the fine-grained semantics of labels at different hierarchy levels. Existing methods often model hierarchy through prediction constraints or representations of individual labels and local relations.  Structural and semantic supervision, however, are usually formulated separately. We instead use complete root-to-node label paths as a shared unit for both. Our framework introduces two auxiliary training objectives while retaining a standard multi-label classifier at inference. PathSimNCE aligns document representations with hierarchical paths using hierarchy-aware contrastive targets. These targets provide graded supervision for structurally related paths while preserving explicit supervision for the ground-truth paths. The complementary semantic objective uses LLM-generated, document-conditioned descriptions of the same positive paths. It thereby provides context-specific label semantics beyond static label representations. Experiments on three benchmarks show consistent improvements: structural alignment is particularly beneficial in deeper hierarchies, while semantic alignment provides additional gains for fine-grained labels. 

\end{abstract}


\section{Introduction}

\begin{figure}[t]
\centering
\includegraphics[width=\columnwidth]{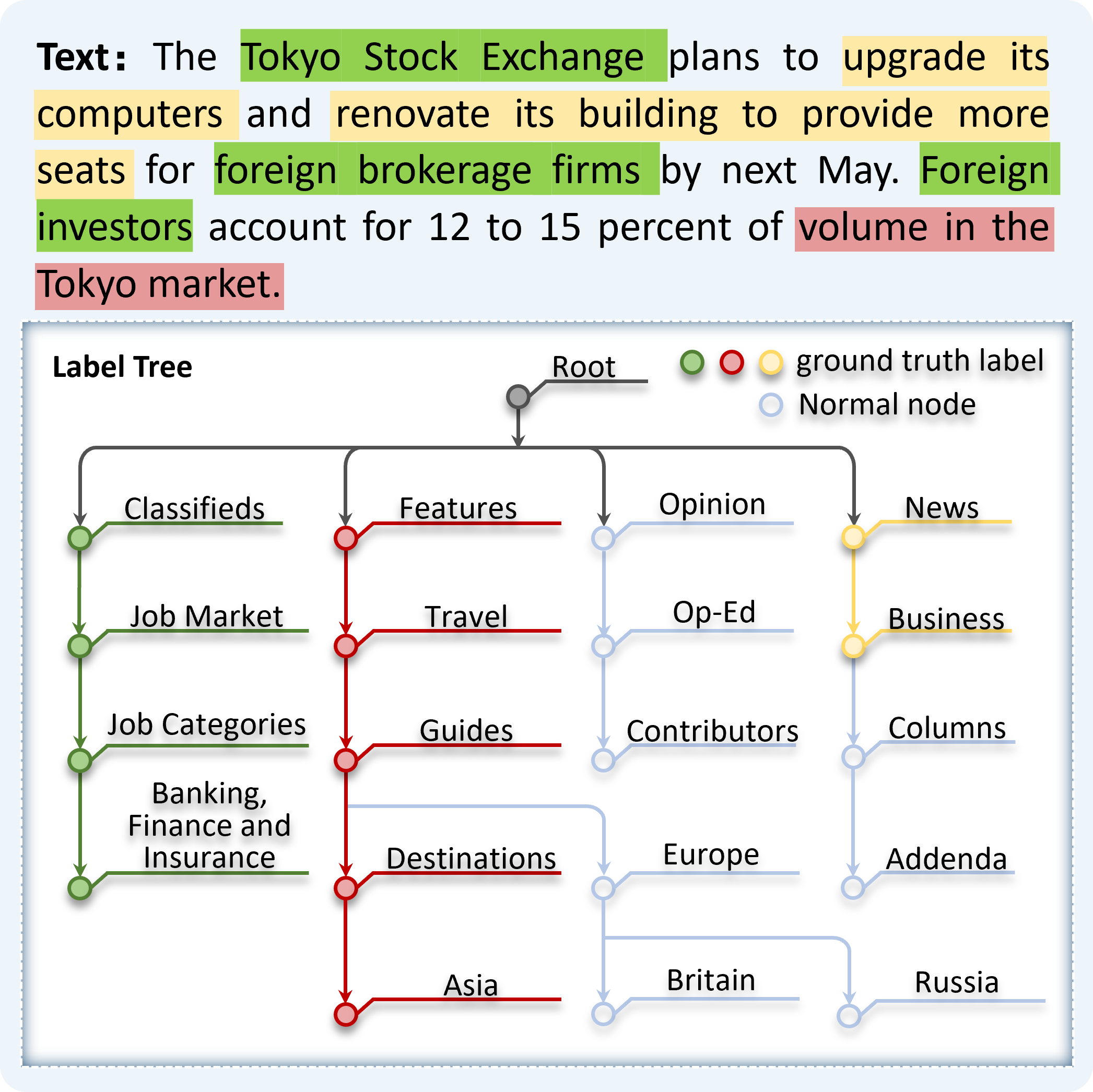}
\caption{An example of HMTC. Labels are organized in a top-down hierarchy from coarse to fine. Labels sharing the same color correspond to nodes along the same hierarchical label path.}
\label{fig:1}
\end{figure}

Hierarchical multi-label text classification (HMTC) assigns multiple labels organized in a predefined hierarchy to a document \citep{liu2023recentadvanceshierarchicalmultilabel} (Figure~\ref{fig:1}). It arises in applications such as news topic classification \citep{info10040150}, scientific publication categorization \citep{chen-etal-2024-retrieval-style}, and book genre annotation \citep{zhou-etal-2020-hierarchy}. Unlike flat multi-label classification, HMTC must account for dependencies among labels at different levels of granularity, ranging from broad categories to fine-grained concepts \citep{11516116}.

Two challenges are particularly important in this setting. First, standard label-wise objectives treat labels largely as independent prediction targets and do not explicitly capture structural relations such as ancestry and siblinghood \citep{zhu-etal-2023-hitin}. This limitation becomes more consequential in deep hierarchies, where closely related labels may share substantial hierarchical context but differ only at lower levels. Second, label names alone often provide insufficient semantic information for distinguishing fine-grained categories \citep{chen-etal-2021-hierarchy,baldini-soares-etal-2019-matching}. Effective HMTC therefore requires supervision that reflects both the structural proximity among labels and their document-specific semantic meanings.

Existing work incorporates hierarchical information primarily through prediction-level constraints or hierarchy-aware representation learning. Prediction-level approaches use structured inference, top-down decoding, or specialized prediction modules to enforce parent--child consistency \citep{kim-etal-2024-hierarchy,10.1007/978-3-031-78107-0_17}. Representation-based approaches encode hierarchical relations using graph encoders, hierarchical attention, or structured label embeddings \citep{sadat-caragea-2022-hierarchical,zhu-etal-2024-hill,zhang-etal-2026-lgsa}. Other methods employ re-weighting, sampling, or metric-learning strategies to improve the prediction of fine-grained labels \citep{NEURIPS2019_621461af,huang-etal-2021-balancing}.  Although effective, these methods often model structure and label semantics separately, leaving complete root-to-node paths underexplored as a common supervision unit for both. 

In this work, we propose a path-level representation learning framework with complementary structural and semantic objectives. Each non-root label is associated with its root-to-node path, which preserves its full ancestral context. Our structural objective, PathSimNCE, aligns document and path representations using hierarchy-aware contrastive targets. It assigns graded supervision to structurally related paths while preserving explicit supervision for the ground-truth paths. Our semantic objective aligns documents with LLM-generated, document-conditioned descriptions of the same positive paths, providing fine-grained semantic supervision beyond static label names. Both objectives are used only during training and complement the standard label-wise classification loss. At inference time, only the text encoder and multi-label classifier are retained.

Our contributions are summarized as follows:
\begin{itemize}
    \item We propose a path-level framework that uses root-to-node paths as a common unit for structural and semantic supervision, without introducing additional inference-time components.
    \item We introduce PathSimNCE, a structural alignment objective that uses hierarchy-aware contrastive targets to provide graded supervision over hierarchical paths while preserving explicit supervision for the ground-truth paths.
    \item We introduce a complementary semantic alignment objective based on document-conditioned path descriptions generated offline by an LLM, providing fine-grained semantic supervision.
   \item We conduct experiments on three HMTC 
   benchmarks and analyze the proposed objectives across hierarchy depths, label frequencies, and representation spaces. 
\end{itemize}

\section{Related Work}

We review prior work on HMTC, focusing on prediction- and constraint-based methods, representation-based hierarchical modeling, and hierarchy-aware contrastive learning.

\textbf{Prediction- and constraint-based approaches} incorporate hierarchical dependencies directly into label prediction. Some methods extend flat multi-label classifiers with hierarchy-regularized objectives or structured inference, while others employ level- or node-specific classifiers followed by top-down or node-wise prediction~\citep{8260658,shimura-etal-2018-hft}. To reduce the complexity of local classification, the hierarchy may be partitioned by level, parent, or node~\citep{STEIN2019216,10.1145/3178876.3186005,krendzelak2021hierarchical}. 
Sequential formulations further represent hierarchical classification as a sequence of label-assignment decisions~\citep{mao-etal-2019-hierarchical, Im_2023}. These approaches can explicitly model prediction dependencies and enforce hierarchical consistency. However, top-down and sequential decisions may propagate errors from coarse to fine levels, and many such methods require hierarchy-specific prediction or inference procedures.

\textbf{Representation-based methods} incorporate hierarchical structure into document or label representations rather than enforcing it solely through prediction-time constraints. Common strategies include graph-based encoders, hierarchy-aware attention, structured message passing, and specialized geometric spaces. For instance, \citet{zhou-etal-2020-hierarchy} encode hierarchical relationships within a unified classification model, while \citet{chen-etal-2021-hierarchy} formulate hierarchical classification as text--label semantic matching. Other studies enrich label representations using knowledge graphs and label attention \citep{liu-etal-2023-enhancing,kumar-toshniwal-2025-hyilr}, or employ tree-isomorphic networks to represent complex hierarchical topologies \citep{zhu-etal-2023-hitin,11036586}. These methods demonstrate the value of learning hierarchy-aware representations, although many introduce dedicated graph, attention, or geometric modules that remain part of the model architecture.

\textbf{Hierarchy-aware contrastive learning} aligns texts, instances, or labels in a shared representation space while using hierarchical relations to construct contrastive supervision. Prior work has explored hierarchy-aware positive and negative sample construction \citep{Xiong2026SCHKHTCSC}, including negative paths or labels selected according to their positions in the hierarchy \citep{zhou-etal-2025-novel}. Other approaches inject hierarchical information into the text encoder \citep{wang-etal-2022-incorporating}, or jointly contrast instances and labels while accounting for their hierarchical relations \citep{yu-etal-2023-instances, 11230573}. Hierarchical margins and rebalancing strategies have also been introduced to improve contrastive learning for infrequent or fine-grained labels \citep{kim-etal-2024-hierarchy,ZHANG2024112153}. Collectively, these studies show that hierarchical relations can improve representation learning, but many contrastive formulations express supervision through discrete positive--negative assignments, margin constraints, or specialized hierarchy-aware modules.

Our framework follows this representation-learning direction but uses root-to-node label paths as a common unit for structural and semantic supervision. PathSimNCE incorporates hierarchical structure into the learning objective through path-similarity-based contrastive targets, assigning graded supervision to structurally related paths while preserving explicit supervision for the ground-truth paths. A complementary description alignment objective supplies document-conditioned semantic supervision for the same paths. Both objectives are used only during training, allowing the model to retain a standard multi-label prediction architecture at inference.

\section{Methodology}
\begin{figure*}[!t]
    \centering
    \includegraphics[width=\textwidth]{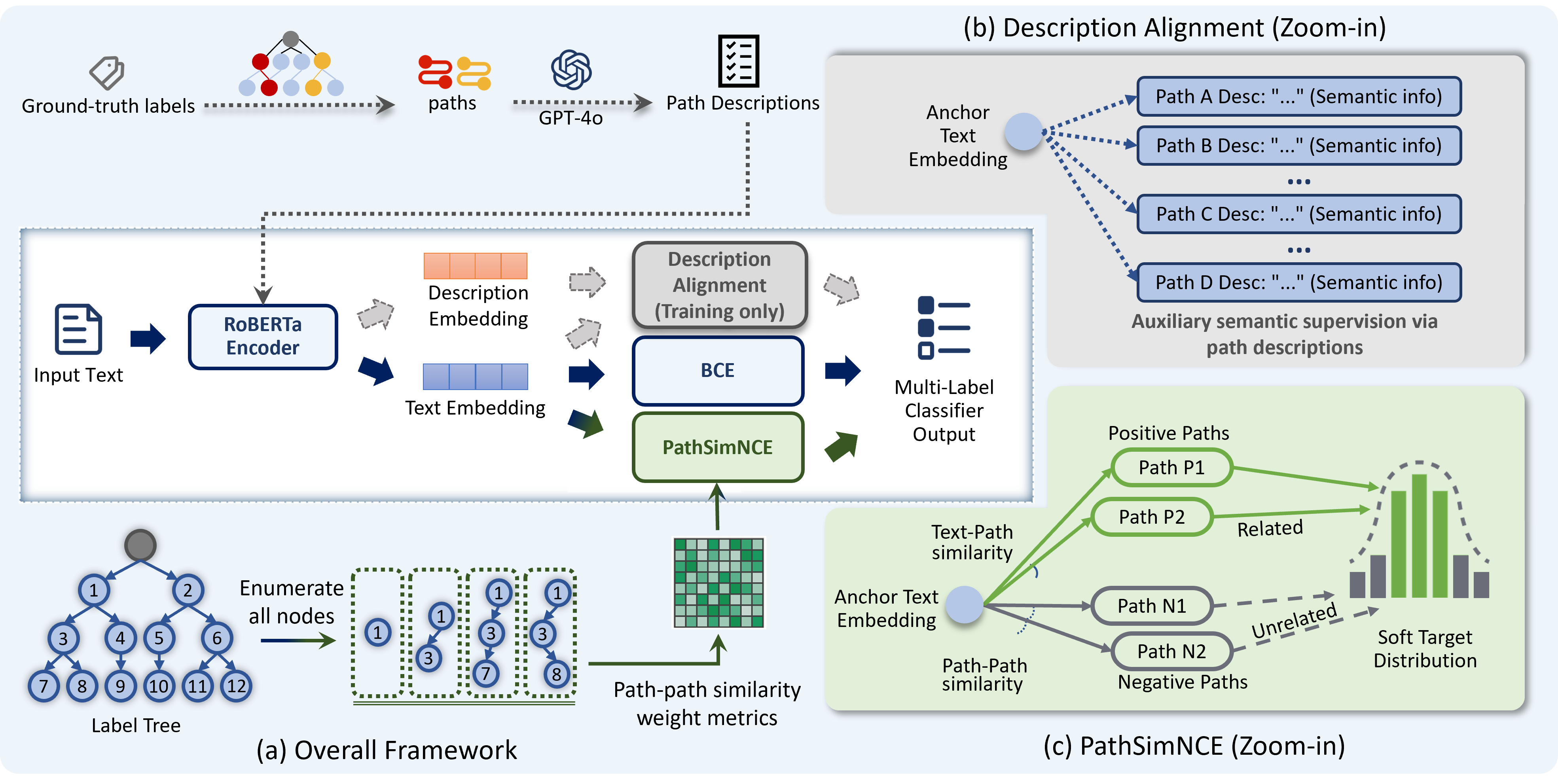}
    \caption{Overview of the proposed hierarchy-aware representation learning framework. (a) Instead of treating labels as isolated identifiers, each label is associated with its root-to-node path in the hierarchy. The model is jointly optimized with three objectives: binary cross-entropy (BCE), PathSimNCE, and description alignment. (b) Description alignment uses LLM-generated path descriptions to provide complementary semantic supervision during training only. (c) PathSimNCE performs hierarchy-aware contrastive learning by combining text--path and path--path similarities to generate soft structural supervision targets.}
    \label{fig:2}  
\end{figure*}

We propose a path-level representation learning framework that augments a standard HMTC classifier with complementary structural and semantic alignment objectives during training (Figure~\ref{fig:2}). For structural alignment, document representations and root-to-node label paths are projected into a shared contrastive space, where PathSimNCE provides graded supervision based on hierarchical path similarity while preserving explicit supervision for the ground-truth paths. For semantic alignment, document representations are aligned with document-conditioned path descriptions generated offline by an LLM. The two alignment objectives are jointly optimized with the standard label-wise classification loss, enabling the model to integrate label prediction, hierarchical structure, and fine-grained semantics during training. At inference time, only the text encoder and standard multi-label classifier are retained, without requiring path representations, descriptions, or a hierarchy-specific decoder.

\subsection{Task Formulation}

We consider hierarchical multi-label text classification over a label set $\mathcal{Y}$ organized in a hierarchy, where each label has at most one parent. Given a document $x$, the goal is to predict its ground-truth label set $\mathcal{Y}_x \subseteq \mathcal{Y}$, which may contain labels at multiple hierarchy levels, including ancestor labels.

For each non-root label $y \in \mathcal{Y}$, we construct a unique hierarchical path by tracing the parent links from $y$ to the root and arranging the resulting nodes in root-to-label order:
\[
p(y)=\bigl(y^{(0)},y^{(1)},\ldots,y^{(d_y)}\bigr),
\]
where $y^{(0)}$ is the root of the corresponding hierarchy branch and $y^{(d_y)}=y$. A root label does not have an associated path because it has no parent. The complete path inventory of a benchmark is denoted as $\mathcal{P}$, and the positive path set associated with $x$ is defined as:
\[
\mathcal{P}_x= \{p(y)\mid y\in\mathcal{Y}_x,\ \operatorname{parent}(y)\text{ exists}\}.
\]
Because a document may have multiple labels, $\mathcal{P}_x$ may contain multiple and potentially overlapping paths. These paths are used only to provide auxiliary supervision during training. The model itself predicts the complete label set $\mathcal{Y}_x$, including root and ancestor labels, using a standard multi-label classifier.

\subsection{Document, Path, and Description Representations}
We map documents, hierarchical paths, and document-conditioned path descriptions into two task-specific representation spaces: a structural alignment space for PathSimNCE and a semantic alignment space for description alignment.

\textbf{Document Representations.} 
Given a document $x$, we obtain its contextualized representation using a RoBERTa~\citep{liu2019robertarobustlyoptimizedbert} encoder:
\begin{equation}
\mathbf{h}_x
=
\operatorname{Pool}\bigl(\operatorname{RoBERTa}(x)\bigr),
\end{equation}
where $\operatorname{Pool}(\cdot)$ denotes the pooling operation applied to the encoder output. We then project $\mathbf{h}_x$ into the structural and semantic alignment spaces:
\begin{equation}
\mathbf{z}_x
=
f_{\mathrm{text}}(\mathbf{h}_x),
\qquad
\mathbf{z}^{\mathrm{desc}}_x
=
f_{\mathrm{desc}}(\mathbf{h}_x).
\end{equation}

\textbf{Path Representations.} 
Each hierarchical path $p\in\mathcal{P}$ is associated with a learnable embedding $\mathbf{h}^{\mathrm{path}}_p\in\mathbb{R}^{d}$ from a trainable embedding table. The path embedding is projected into the structural alignment space:
\begin{equation}
\mathbf{z}^{\mathrm{path}}_p
=
f_{\mathrm{path}}\bigl(\mathbf{h}^{\mathrm{path}}_p\bigr).
\end{equation}
These path embeddings are jointly optimized with the text encoder and serve as structural anchors in PathSimNCE.

\textbf{Description Representations.} 
For each ground-truth document--path pair $(x,p)$, where $p\in\mathcal{P}_x$, we use an LLM to generate a document-conditioned path description $d_{x,p}$. The description explains how the content of $x$ corresponds to the hierarchical labels represented by $p$ and is used only during training. We encode it using the same RoBERTa encoder and project it into the semantic alignment space:
\begin{equation}
\begin{aligned}
\mathbf{h}^{\mathrm{desc}}_{x,p}
&=
\operatorname{Pool}\bigl(\operatorname{RoBERTa}(d_{x,p})\bigr), \\
\mathbf{z}^{\mathrm{desc}}_{x,p}
&=
f_{\mathrm{desc}}\bigl(\mathbf{h}^{\mathrm{desc}}_{x,p}\bigr).
\end{aligned}
\end{equation}
The resulting description representations provide auxiliary semantic views of the document--path relationships, complementing the structural supervision supplied by PathSimNCE.

\subsection{PathSimNCE: Path Similarity-Aware InfoNCE}

We propose PathSimNCE, a hierarchy-aware contrastive objective that aligns document representations with hierarchical paths while explicitly modeling structural similarities among paths. For each document $x_i$ with positive path set $\mathcal{P}_{x_i}$, we construct a sample-specific candidate set $\mathcal{C}_i$ from the global path inventory $\mathcal{P}$. Document--path similarities define a predicted distribution over $\mathcal{C}_i$, while hierarchical relationships among paths are used to construct a soft target distribution.

\paragraph{Hierarchical Path Similarity.}
For two paths $p,p'\in\mathcal{P}$, we define their hierarchical
distance based on their lowest common ancestor (LCA):
\begin{equation}
\label{eq:path_distance}
\operatorname{dist}(p,p')
=
|p|+|p'|
-
2\left|\operatorname{LCA}(p,p')\right|,
\end{equation}
where $|\cdot|$ denotes the number of nodes. 
We convert this distance into a structural similarity score:
\begin{equation}
\label{eq:path_similarity}
\operatorname{sim}_{\mathrm{path}}(p,p')
=
\exp\left(
-\lambda\,\operatorname{dist}(p,p')
\right),
\end{equation}
where $\lambda>0$ controls the decay rate.

\paragraph{Candidate Path Construction and Negative Path Sampling.}
For document $x_i$, we first remove its positive paths $\mathcal{P}_{x_i}$ from the complete path inventory to form the negative path pool:
\begin{equation}
\label{eq:negative_path_pool}
\mathcal{U}_i
=
\mathcal{P}\setminus\mathcal{P}_{x_i}.
\end{equation}

For each candidate path $p\in\mathcal{U}_i$, we define its distance from the positive path set as the minimum hierarchical distance to any
positive path:
\begin{equation}
\label{eq:candidate_positive_distance}
\delta_i(p)
=
\min_{p^+\in\mathcal{P}_{x_i}}
\operatorname{dist}(p,p^+),
\end{equation}
where $p^+$ denotes a positive path associated with $x_i$. We then group candidate paths according to this distance:
\begin{equation}
\label{eq:distance_group}
\mathcal{G}_{i,r}
=
\left\{
p\in\mathcal{U}_i
\mid
\delta_i(p)=r
\right\},
\end{equation}
where $\mathcal{G}_{i,r}$ contains all candidate paths whose minimum distance to the positive path set is $r$. A smaller value of $r$ indicates that the paths in the group are hierarchically closer to the ground-truth paths.

Within each distance group, we rank candidate paths according to the current model predictions. Specifically, the prediction score of a path is defined as the average predicted probability of its constituent labels:
\begin{equation}
\label{eq:path_prediction_score}
\operatorname{score}_i(p)
=
\frac{1}{|p|}
\sum_{y\in p}
\hat{y}_{i},
\end{equation}
where $\hat{y}_{i}$ is the probability assigned by the 
classifier to label $y$ for document $x_i$. Paths within each
$\mathcal{G}_{i,r}$ are sorted by
$\operatorname{score}_i(p)$ in descending order.

We sample negative paths in rounds, visiting the distance groups from nearest to farthest and selecting the highest-ranked remaining path
from each group in every round. Sampling continues until $K$ negative paths have been selected, yielding the negative-path set $\mathcal{N}_i$.

Finally, the sample-specific candidate set is: 
\begin{equation}
\label{eq:candidate_path_set}
\mathcal{C}_i
=
\mathcal{P}_{x_i}\cup\mathcal{N}_i.
\end{equation}
This sampling strategy prioritizes hierarchically close and high-scoring negative paths while retaining candidates from different
distance groups.

\paragraph{PathSimNCE Objective.}

Given the candidate set $\mathcal{C}_i$, we compute the 
cosine similarity between document $x_i$ and each candidate path
$p\in\mathcal{C}_i$:
\begin{equation}
\label{eq:text_path_similarity}
a_i(p)
=
\frac{
\mathbf{z}_{x_i}^{\top}\mathbf{z}^{\mathrm{path}}_p
}{
\tau_{\mathrm{h}}
\|\mathbf{z}_{x_i}\|_2
\|\mathbf{z}^{\mathrm{path}}_p\|_2
},
\end{equation}
where $\tau_{\mathrm{h}}>0$ is a temperature hyperparameter. These similarity scores define the predicted distribution over the candidate paths:
\begin{equation}
\label{eq:predicted_path_distribution}
\hat{q}_i(p)
=
\frac{
\exp\bigl(a_i(p)\bigr)
}{
\displaystyle
\sum_{\tilde{p}\in\mathcal{C}_i}
\exp\bigl(a_i(\tilde{p})\bigr)
}.
\end{equation}

To construct the target distribution, we first aggregate the structural
similarity of each candidate path to all positive paths:
\begin{equation}
\label{eq:structural_target_score}
s_i(p)
=
\sum_{p^+\in\mathcal{P}_{x_i}}
\operatorname{sim}_{\mathrm{path}}(p,p^+),
\end{equation}
We then normalize these scores over the candidate set to obtain the structure-aware distribution:
\begin{equation}
\label{eq:structural_target_distribution}
q_i^{s}(p)
=
\frac{
s_i(p)
}{
\displaystyle
\sum_{\tilde{p}\in\mathcal{C}_i}
s_i(\tilde{p})
}.
\end{equation}

To retain explicit supervision from the ground-truth paths, we additionally define a uniform positive distribution:
\begin{equation}
\label{eq:uniform_positive_distribution}
q_i^{\mathrm{gt}}(p)
=
\begin{cases}
\displaystyle
\frac{1}{|\mathcal{P}_{x_i}|},
& p\in\mathcal{P}_{x_i},\\[6pt]
0,
& p\notin\mathcal{P}_{x_i}.
\end{cases}
\end{equation}

The final soft target is obtained by interpolating the two distributions:
\begin{equation}
\label{eq:pathsimnce_target}
q_i(p)
=
(1-\gamma)q_i^{s}(p)
+
\gamma q_i^{\mathrm{gt}}(p),
\end{equation}
where $\gamma\in[0,1]$ controls the contribution of exact ground-truth supervision. 

For a mini-batch $\{x_i\}_{i=1}^{B}$, PathSimNCE is defined as the cross-entropy between the target distribution $q_i$ and the predicted distribution $\hat{q}_i$:
\begin{equation}
\label{eq:pathsimnce_objective}
\mathcal{L}_{\mathrm{PathSimNCE}}
=
-\frac{1}{B}
\sum_{i=1}^{B}
\sum_{p\in\mathcal{C}_i}
q_i(p)\log\hat{q}_i(p).
\end{equation}
This objective aligns document representations with their positive paths while propagating graded supervision to structurally related candidate paths.

\subsection{Description Alignment}

While PathSimNCE captures hierarchical structure through path-level supervision, it does not explicitly model the semantic correspondence
between a document and its positive paths. To provide complementary semantic supervision, we introduce a description alignment objective that aligns each document with the LLM-generated descriptions of its positive paths.

For a mini-batch $\{x_i\}_{i=1}^{B}$, the description alignment loss is defined as follows:
\begin{equation}
\label{eq:description_alignment_loss}
\resizebox{0.98\columnwidth}{!}{$\displaystyle
\mathcal{L}_{\mathrm{DESC}}
=
-\frac{1}{B}
\sum_{i=1}^{B}
\frac{1}{|\mathcal{P}_{x_i}|}
\sum_{p\in\mathcal{P}_{x_i}}
\log
\frac{
\exp\left(
(\mathbf{z}^{\mathrm{desc}}_{x_i})^{\top}
\mathbf{z}^{\mathrm{desc}}_{x_i,p}
/\tau_{\mathrm{desc}}
\right)
}{
\displaystyle
\sum_{j=1}^{B}
\sum_{\tilde p\in\mathcal{P}_{x_j}}
\exp\left(
(\mathbf{z}^{\mathrm{desc}}_{x_i})^{\top}
\mathbf{z}^{\mathrm{desc}}_{x_j,\tilde p}
/\tau_{\mathrm{desc}}
\right)
}
.$}
\end{equation}

Here, $\tau_{\mathrm{desc}}>0$ is a temperature hyperparameter. The numerator measures the similarity between document $x_i$ and the
description associated with one of its positive paths $p\in\mathcal{P}_{x_i}$, while the denominator contrasts $x_i$ against
all path descriptions in the mini-batch. By averaging over all $p\in\mathcal{P}_{x_i}$, the objective aligns the document representation with the semantic descriptions of all its positive
paths.

\begin{table*}[t]
\centering
{\small
\begin{tabular*}{\textwidth}{@{\extracolsep{\fill}}lcccccc@{}}
\toprule
\multirow{2}{*}{\textbf{Model}}
& \multicolumn{2}{c}{\textbf{WOS}}
& \multicolumn{2}{c}{\textbf{BGC}}
& \multicolumn{2}{c}{\textbf{NYT}} \\

\cmidrule(lr){2-3}
\cmidrule(lr){4-5}
\cmidrule(l){6-7}

& \textbf{Micro-F1} & \textbf{Macro-F1}
& \textbf{Micro-F1} & \textbf{Macro-F1}
& \textbf{Micro-F1} & \textbf{Macro-F1} \\
\midrule

RoBERTa-base
& 86.45
& 78.89
& 81.40
& 67.13
& 81.79
& 72.06 \\

HGCLR \citep{wang-etal-2022-incorporating}
& 86.90 & 80.98 & 81.24 & 67.00 & 81.08 & 71.45 \\

HiTIN \citep{zhu-etal-2023-hitin}
& 86.63 & 80.92 & 80.45 & 66.22 & 80.30 & 69.80 \\

HILL \citep{zhu-etal-2024-hill}
& 87.35 & 82.04 & 81.15 & 66.03 & 80.96 & 70.21 \\

HALB \citep{ZHANG2024112153}
& 86.97 & 81.26 & 81.34 & 67.00 & 80.70 & 70.93 \\

HYDRA Local \citep{karl-scherp-2025-hydra}
& 86.90 & 81.14 & 82.18 & 66.01 & 81.87 & 72.43 \\

DFG \citep{LIU2025104065}
& 87.44 & 81.80 & 81.20 & 66.73 & 81.28 & 72.15 \\

HDCL-SRC \citep{11462716}
& 87.03 & 81.47 & 81.56 & 67.92 & 81.69 & 72.71 \\

LGSA \citep{zhang-etal-2026-lgsa}
& 88.11 & \textbf{82.93}
& 82.12 & 68.40
& 81.76 & 69.95 \\

\midrule

Ours
& \textbf{88.15}{\scriptsize\kern-0.08em$\pm$0.12}
& 81.23{\scriptsize\kern-0.08em$\pm$0.27}
& \textbf{82.39}{\scriptsize\kern-0.08em$\pm$0.15}
& \textbf{68.85}{\scriptsize\kern-0.08em$\pm$0.33}
& \textbf{82.52}{\scriptsize\kern-0.08em$\pm$0.22}
& \textbf{73.18}{\scriptsize\kern-0.08em$\pm$0.26} \\

\bottomrule
\end{tabular*}
}
\caption{Performance comparison with previous methods on the WOS, BGC, and NYT datasets. To ensure a fair comparison, all reproduced methods use the same RoBERTa-base encoder, regardless of the encoders originally adopted in their respective papers. Results for our method are reported as mean $\pm$ standard deviation over three runs. Bold indicates the best result in each column. Except for HYDRA Local, whose results are taken from the original paper, all reported results are obtained from our own experiments.}
\label{tab:2}
\end{table*}

\begin{table}[t]
\centering
\small
\begin{tabular}{@{}lcccc@{}}
\toprule
            \textbf{Dataset} & \textbf{\#Labels} & \textbf{Depth} & \textbf{Avg \# label} & \textbf{Avg length}\\ 
\midrule
\textbf{WOS} & 141                & 2              & 2.00                & 207.62      \\ 
\textbf{BGC} & 146                & 4              & 3.01                & 162.88      \\ 
\textbf{NYT} & 166                & 8              & 7.58                & 605.90      \\
 
\bottomrule
\end{tabular}
\caption{Statistics of the datasets.} 
\label{tab:1}
\end{table}

\subsection{Joint Learning}

The proposed framework is trained end-to-end by jointly optimizing the label prediction objective and the two auxiliary representation learning objectives. The standard multi-label classifier is trained using the binary cross-entropy loss:
\begin{equation}
    \mathcal{L}_{\text{BCE}} = -\frac{1}{|\mathcal{Y}|}\sum_{y \in \mathcal{Y}}\Big[y_i \log \hat{y}_{i} + (1 - y_i)\log(1 - \hat{y}_{i})\Big],
\end{equation}
where $y_{i}\in\{0,1\}$ denotes the ground-truth indicator of label $y$ for input $x_i$, and $\hat y_{i}$ denotes the corresponding predicted probability. 

The overall training objective is:
\begin{equation}
\label{eq:overall_objective}
\mathcal{L}
=
\mathcal{L}_{\mathrm{BCE}}
+
\beta\,\mathcal{L}_{\mathrm{PathSimNCE}}
+
\beta_{\mathrm{desc}}\,\mathcal{L}_{\mathrm{DESC}},
\end{equation}
where $\beta$ and $\beta_{\mathrm{desc}}$ control the contributions of PathSimNCE and the description alignment objective, respectively.

The three objectives provide complementary supervision:
$\mathcal{L}_{\mathrm{BCE}}$ directly supervises label prediction,
$\mathcal{L}_{\mathrm{PathSimNCE}}$ incorporates hierarchical
structure through path-level alignment, and
$\mathcal{L}_{\mathrm{DESC}}$ provides document-conditioned semantic supervision. At inference time, the model retains only the text encoder and the standard multi-label classifier, with no additional inputs or computation.

\section{Experimental settings}

We evaluate our method on three widely used HMTC benchmarks: WOS-46985 (WOS) \citep{8260658}, which contains scientific abstracts; the Blurb Genre Collection (BGC) \citep{aly-etal-2019-hierarchical},
which contains multi-genre book blurbs; and the New York Times Annotated Corpus (NYT) \citep{ACQUIRED}, which contains hierarchically labeled news articles. Their label hierarchies have maximum depths of 2, 4, and 8, respectively. Dataset statistics are summarized in Table~\ref{tab:1}.

We implement all models in PyTorch \citep{NEURIPS2019_bdbca288} and train them on a single NVIDIA H100 80GB GPU. 
We use AdamW and set the RoBERTa learning rate $lr$ to $3.5\times10^{-5}$ and the task-specific learning rate $head\_lr$ to $1\times10^{-4}$. Models are trained with a batch size of 32 for up to 50 epochs, with early stopping on validation micro-F1 using a patience of 6 epochs. We set $\tau_{\mathrm{h}}=\tau_{\mathrm{desc}}=0.07$,
$\beta=0.2$, $\beta_{\mathrm{desc}}=0.1$, and $\gamma=0.1$.
The hyperparameter search space and tuning results are provided in Appendix~\ref{sec:hyperparameter_tuning}. Path descriptions are
generated offline using GPT-4o \citep{openai_gpt4o}; prompting details are provided in
Appendix~\ref{sec:supp_prompt}.
Following prior work, we report micro-F1 and macro-F1, using a probability threshold of 0.5 to obtain label predictions. Our code is available at \url{https://anonymous.4open.science/r/jubilant-octo-journey-D685/}.


\section{Results and Discussion}

\subsection{Main Results}

Table~\ref{tab:2} compares our method with representative HMTC baselines on WOS, BGC, and NYT. All methods use RoBERTa-base as the
text encoder under the same experimental setting. Our method achieves the best Micro-F1 on all three datasets and the best Macro-F1 on BGC and NYT.

On WOS, our method obtains a Micro-F1 score of $88.15$, slightly surpassing the strongest baseline by $0.04$ points. Its Macro-F1 score
of $81.23$ remains competitive, although it is lower than the best baseline result of $82.93$. This may be because WOS has a relatively shallow hierarchy, offering fewer structural relationships for path-level supervision to exploit. On BGC, our method achieves $82.39$
Micro-F1 and $68.85$ Macro-F1, improving over the corresponding strongest baselines by $0.21$ and $0.45$ points, respectively. The
largest Micro-F1 improvement is observed on NYT, where our method reaches $82.52$ Micro-F1 and $73.18$ Macro-F1, exceeding the strongest
baselines by $0.65$ and $0.47$ points.

These results demonstrate the effectiveness of our framework across hierarchies of different depths, while requiring only the standard
multi-label classification branch at inference time. Representative case studies illustrating the effects of structural and semantic supervision are provided in
Appendix~\ref{sec:case_analysis}.

\begin{table*}[!t]
\centering
{
\small
\setlength{\tabcolsep}{2.5pt}

\newcommand{\scorestd}[2]{%
\makebox[2.8em][r]{#1}%
\makebox[2.3em][l]{{\scriptsize\kern-0.08em$\pm$#2}}%
}

\begin{tabular*}{0.98\textwidth}{@{\extracolsep{\fill}}lcccccc@{}}
\toprule
& \multicolumn{2}{c}{\textbf{WOS}} 
& \multicolumn{2}{c}{\textbf{BGC}} 
& \multicolumn{2}{c}{\textbf{NYT}} \\

\cmidrule(lr){2-3}
\cmidrule(lr){4-5}
\cmidrule(lr){6-7}

\textbf{Model}
& \textbf{Micro-F1} & \textbf{Macro-F1}
& \textbf{Micro-F1} & \textbf{Macro-F1}
& \textbf{Micro-F1} & \textbf{Macro-F1} \\
\midrule

BCE
& \scorestd{86.45}{0.16}
& \scorestd{78.89}{0.18}
& \scorestd{81.40}{0.13}
& \scorestd{67.13}{0.30}
& \scorestd{81.79}{0.18}
& \scorestd{72.06}{0.22} \\

BCE+PathSimNCE
& \scorestd{87.53}{0.23}
& \scorestd{80.66}{0.33}
& \scorestd{81.68}{0.19}
& \scorestd{67.85}{0.29}
& \scorestd{82.29}{0.12}
& \scorestd{72.94}{0.31} \\

BCE+DESC
& \scorestd{87.11}{0.22}
& \scorestd{79.68}{0.32}
& \scorestd{81.60}{0.15}
& \scorestd{67.73}{0.30}
& \scorestd{82.14}{0.21}
& \scorestd{72.43}{0.33} \\

\midrule

No Description
& \scorestd{87.53}{0.23}
& \scorestd{80.66}{0.33}
& \scorestd{81.68}{0.19}
& \scorestd{67.85}{0.29}
& \scorestd{82.29}{0.12}
& \scorestd{72.94}{0.31} \\

Template Description
& \scorestd{86.17}{0.14}
& \scorestd{77.76}{0.22}
& \scorestd{81.72}{0.19}
& \scorestd{67.90}{0.29}
& \scorestd{82.32}{0.13}
& \scorestd{72.98}{0.36} \\

LLM-generated Description
& \scorestd{88.15}{0.12}
& \scorestd{81.23}{0.27}
& \scorestd{82.39}{0.15}
& \scorestd{68.85}{0.33}
& \scorestd{82.52}{0.22}
& \scorestd{73.18}{0.26} \\

\bottomrule 
\end{tabular*}
}

\caption{Ablation results on the three datasets. The upper block examines the contributions of the training objectives, and the lower block compares different forms of semantic supervision while retaining BCE and PathSimNCE. Results are reported as mean $\pm$ standard deviation over three runs.}
\label{tab:3}
\end{table*}

\begin{figure}[t]
    \includegraphics[width=\columnwidth,keepaspectratio]{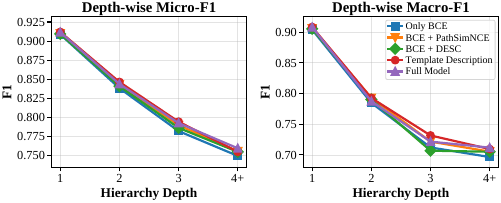}
    \caption{Depth-wise Micro-F1 and Macro-F1 on NYT. Labels at depths of four or greater are grouped into the 4+ category because of data sparsity. The virtual \textsc{Top} label at depth 0 is excluded.}
    \label{fig:depth}  
\end{figure}

\subsection{Ablation Studies}

We examine the contributions of the two auxiliary objectives and the effect of different descriptions in Table~\ref{tab:3}. All variants use the same encoder, classifier, and training configuration.

\textbf{Contributions of Training Objectives.}
Starting from the BCE baseline, we separately add PathSimNCE and description alignment. Both objectives improve Micro-F1 and Macro-F1
on all three datasets. PathSimNCE produces larger gains, improving Micro-F1 by $1.08$, $0.28$, and $0.50$ points on WOS, BGC, and NYT,
respectively. Description alignment also consistently improves the BCE baseline, although its gains are smaller when used alone.
Combining the two objectives yields the best overall results, outperforming BCE+PathSimNCE by $0.62$, $0.71$, and $0.23$ Micro-F1 points on the three datasets. These results show that structural path supervision provides the main improvement, while description alignment offers complementary semantic supervision.

\textbf{Effects of Description Variants.}
We next fix BCE+PathSimNCE and vary the semantic supervision used for description alignment. We compare no description alignment,
template-based descriptions constructed from predefined template using label names, and
document-conditioned descriptions generated by GPT-4o. Prompting details and template examples are provided in
Appendix~\ref{sec:supp_prompt_example}. Template-based descriptions provide only marginal improvements on BGC and NYT and substantially degrade performance on WOS. In contrast, LLM-generated descriptions
consistently achieve the best results, improving over the variant
without description alignment by $0.62$, $0.71$, and $0.23$ Micro-F1 points on WOS, BGC, and NYT, respectively. This comparison suggests that semantic alignment benefits from descriptions that capture the document-specific meaning of a path rather than relying on fixed verbalizations of its labels.

\subsection{Performance across Hierarchy Depths}

To examine model behavior at different hierarchy levels, we report depth-wise Micro-F1 and Macro-F1 on NYT in Figure~\ref{fig:depth}. Labels at depths of four or greater are grouped into a single 4+ category because of data sparsity. Both metrics
generally decrease with depth, reflecting the increasing difficulty of predicting fine-grained labels.

Compared with BCE, BCE+PathSimNCE achieves consistent improvements across all depth groups, with the gains remaining evident at depths 3 and 4+. The full model obtains the best Micro-F1 and Macro-F1 in the
4+ group, indicating that semantic supervision further complements structural supervision for the deepest labels. Although template-based descriptions perform competitively at some shallower levels, their overall improvements are less consistent, in line with the aggregate results in Table~\ref{tab:3}. These findings suggest that PathSimNCE provides robust supervision across hierarchy levels, while document-conditioned descriptions are particularly helpful for fine-grained labels. We additionally analyze
model performance across different label-frequency groups in
Appendix~\ref{sec:label_frequency}.

\begin{figure}[!t]
    \centering    \includegraphics[width=\columnwidth,height=0.38\textheight,keepaspectratio]{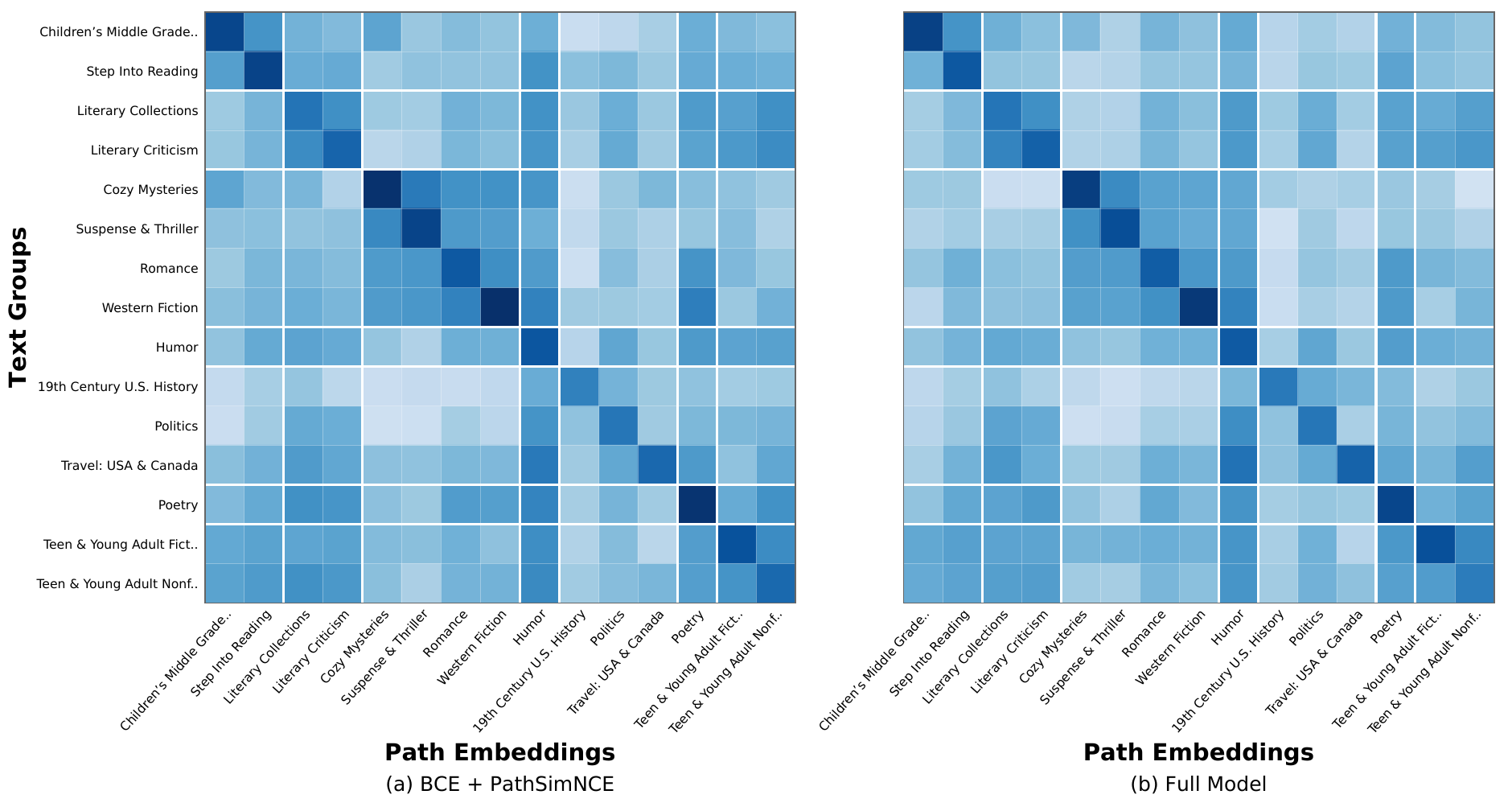}
    \caption{Document--path cosine similarity heatmaps on BGC for (a) BCE+PathSimNCE and (b) the full model. Each row represents a group of test documents sharing the corresponding ground-truth path, and each column represents a selected path embedding. Both panels use the same color scale, and darker colors indicate higher average cosine similarity.}
    \label{fig:4}
\end{figure}

\subsection{Document--Path Alignment Analysis}


To qualitatively examine the structural alignment learned by our model, we visualize document--path cosine similarities on BGC. We select 15 representative paths covering several hierarchy branches and arrange paths from the same branch next to one another where possible. For each selected path, we sample the same number of test documents for which it is a ground-truth path.

As shown in Figure~\ref{fig:4}, both
BCE+PathSimNCE and the full model exhibit clear diagonal dominance, indicating that documents are represented closest to their ground-truth paths. Paths from related hierarchy branches generally show higher local similarities than paths from more distant branches, which is consistent with the graded structural supervision introduced by PathSimNCE. The full model retains this alignment after incorporating description-based semantic supervision. Additional t-SNE visualizations of document and path representations
are provided in Appendix~\ref{sec:tsne_visualization}.

\section{Conclusion}

We propose a path-level structural and semantic alignment framework for hierarchical multi-label text classification. It uses complete root-to-node label paths as a shared unit for structural and semantic supervision. We use two auxiliary training objectives. PathSimNCE aligns document representations with hierarchical paths using graded structural supervision while preserving explicit supervision for the ground-truth paths. The complementary semantic objective uses document-conditioned descriptions of the same positive paths to capture context-specific label semantics. Both objectives are training-only. The model retains a standard multi-label classifier without additional inference-time inputs or computation.

Experiments on three datasets show consistent improvements. Structural alignment is particularly beneficial in deeper hierarchies, while semantic alignment provides additional gains for fine-grained labels. These results support hierarchical paths as a common interface for combining structural and semantic supervision. In future, we plan to extend the framework to more general hierarchies, including taxonomies allowing multiple parents per label. We will also investigate more scalable approaches to constructing and using path descriptions for large label hierarchies.

\section*{Limitations}

Our framework has several limitations. First, it assumes that a reliable label hierarchy is available and that each non-root label has a unique parent. Consequently, the current path construction and structural targets cannot be directly applied to more general taxonomies in which labels have multiple parents. Moreover, when the given hierarchy is noisy, incomplete, or weakly related to label
semantics, path-level supervision may become less informative.

Second, description alignment relies on LLM-generated,
document-conditioned path descriptions. Although these descriptions are generated offline and used only during training, their generation introduces additional pre-processing cost, particularly for large
datasets with many document--path pairs. Inaccurate, overly generic, or biased descriptions may also introduce noisy semantic supervision and reduce the benefits of description alignment.

Finally, the proposed objectives inject structural and semantic information during training but do not impose hard hierarchical constraints on predictions. Because labels are still predicted independently by a standard multi-label classifier, the resulting label set is not guaranteed to be hierarchically consistent at inference time.

\section*{Ethics Statement}
We are using the publicly-available datasets, and we do not see any ethics issues in this paper.

\section*{Usage of Large Language Models}
We use GPT-4o to generate document-conditioned path descriptions for
ground-truth document--path pairs. The descriptions are generated offline and used only as auxiliary semantic supervision during training; GPT-4o is not involved at inference time. In addition, GPT-4o is used to assist manuscript polishing for improving language clarity and readability.

\section*{Impact Statements}
This work studies representation learning for hierarchical multi-label text classification while retaining a standard multi-label classifier at inference time. The proposed framework is evaluated on publicly available benchmark datasets. Its effectiveness depends on the quality of the label hierarchy and, for description alignment, the quality of the LLM-generated path descriptions.

\bibliography{custom, anthology}

\section*{Appendix}

\appendix

\begin{table}[!t]
    \centering

    \small
    \setlength{\tabcolsep}{5.5pt}
    \renewcommand{\arraystretch}{1.12}
    \begin{tabular}{@{}cccc@{}}
        \toprule
        Parameter & Value & Micro-F1 & Macro-F1 \\
        \midrule

        \multirow[c]{3}{*}{$\beta$}
        & 0.3 & 82.06 & 72.70 \\
        & \textbf{0.2} & \textbf{82.52} & \textbf{73.18} \\
        & 0.1 & 82.20 & 72.83 \\

        \midrule

        \multirow[c]{2}{*}{$\beta_{\mathrm{desc}}$}
        & 0.2 & 82.32 & 72.88 \\
        & \textbf{0.1} & \textbf{82.52} & \textbf{73.18} \\

        \midrule

        \multirow[c]{3}{*}{$\tau_h$}
        & 0.10 & 82.30 & 72.66 \\
        & \textbf{0.07} & \textbf{82.52} & \textbf{73.18} \\
        & 0.05 & 82.28 & 72.56 \\

        \midrule

        \multirow[c]{3}{*}{$\tau_{\mathrm{desc}}$}
        & 0.10 & 82.23 & 72.51 \\
        & \textbf{0.07} & \textbf{82.52} & \textbf{73.18} \\
        & 0.05 & 82.31 & 72.48 \\

        \midrule

        \multirow[c]{3}{*}{$\gamma$}
        & 0.2 & 81.37 & 71.67 \\
        & \textbf{0.1} & \textbf{82.52} & \textbf{73.18} \\
        & 0.0 & 82.26 & 72.51 \\

        \midrule

        \multirow[c]{3}{*}{$lr$}
        & $5\times10^{-5}$ & 82.57 & 72.82 \\
        & $\mathbf{3.5\times10^{-5}}$ & \textbf{82.52} & \textbf{73.18} \\
        & $3\times10^{-5}$ & 82.41 & 73.09 \\

        \midrule

        \multirow[c]{3}{*}{$head\_lr$}
        & $2\times10^{-4}$ & 82.36 & 72.92 \\
        & $\mathbf{1\times10^{-4}}$ & \textbf{82.52} & \textbf{73.18} \\
        & $8\times10^{-5}$ & 82.39 & 73.24 \\

        \bottomrule
    \end{tabular}

    \caption{Hyperparameter sensitivity on the NYT dataset. Bold indicates the hyperparameter settings used in the final model and their corresponding performance.}
    \label{tab:nyt_hyperparameters}
\end{table}

\section{Hyperparameter Analysis}
\label{sec:hyperparameter_tuning}

To examine the sensitivity of the proposed framework to its main
hyperparameters, we conduct controlled experiments on the NYT dataset by
varying one parameter at a time while keeping the remaining settings fixed.
We consider the loss weights $\beta$ and $\beta_{\mathrm{desc}}$,
the temperature parameters $\tau_h$ and $\tau_{\mathrm{desc}}$,
the interpolation coefficient $\gamma$, and the learning rates of the
RoBERTa encoder and task-specific parameters. Bold entries in
Table~\ref{tab:nyt_hyperparameters} denote the settings adopted in the
final model. This analysis allows us to examine how each component
influences the overall performance while controlling for the effects
of the remaining hyperparameters.

As shown in Table~\ref{tab:nyt_hyperparameters}, $\beta=0.2$,
$\beta_{\mathrm{desc}}=0.1$, and
$\tau_h=\tau_{\mathrm{desc}}=0.07$ achieve strong overall performance
within the tested ranges. Both temperature parameters exhibit
performance variations when their values are increased or decreased
from 0.07, with the selected settings yielding the highest Micro-F1
and Macro-F1 in their respective experiments. For $\gamma$, the
selected value of 0.1 also yields the best overall results, suggesting
that a moderate balance between structure-aware and explicit
ground-truth supervision is beneficial in this setting.

The learning-rate settings show slight trade-offs between Micro-F1
and Macro-F1. A larger RoBERTa learning rate gives a slightly higher
Micro-F1, while a smaller task-specific learning rate yields a
marginally higher Macro-F1. Nevertheless, the selected values of
$3.5\times10^{-5}$ and $1\times10^{-4}$ provide balanced overall
performance. Overall, the model remains reasonably stable across
the evaluated hyperparameter ranges.

    

\begin{figure}[!b]
    \centering
    \includegraphics[width=\linewidth,keepaspectratio]
    {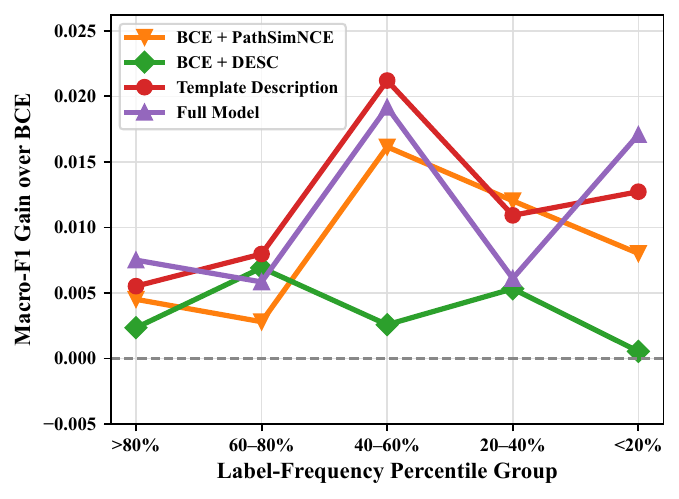}
    \caption{Macro-F1 gains over the BCE baseline across label-frequency percentile groups on NYT. Labels are ranked by their numbers of training instances and divided into five equally sized groups from the most frequent to the least frequent. Positive values indicate improvements over BCE.}
    \label{fig:longtail}
\end{figure}

\begin{table*}[t]
\centering
\small
\setlength{\tabcolsep}{4pt}
\renewcommand{\arraystretch}{1.18}
\setlength{\heavyrulewidth}{\lightrulewidth}
\begin{tabularx}{\textwidth}{
@{}
>{\centering\arraybackslash}m{0.19\textwidth}
>{\raggedright\arraybackslash}X
>{\raggedright\arraybackslash}X
@{}
}
\toprule
\textbf{Item}
& \multicolumn{1}{c}{\textbf{Case 1: \textit{Count Zero}}}
& \multicolumn{1}{c}{\textbf{Case 2: \textit{Coriolanus}}} \\
\midrule

\textbf{Input Text}
&
A stylish, streetsmart, frighteningly probable parable of the
future \ldots{} A corporate mercenary wakes in a reconstructed
body, a beautiful woman by his side. Then Hosaka Corporation
reactivates him, for a mission more dangerous than the one
he's recovering from: to get a defecting chief of R\&D---and
the biochip he's perfected---out intact. \ldots{} But this proves
to be of supreme interest to certain other parties---some of
whom aren't remotely human \ldots{}
&
The legendary Pelican Shakespeare series features authoritative
and meticulously researched texts paired with scholarship by
renowned Shakespeareans. Each book includes an essay on the
theatrical world of Shakespeare's time, an introduction to the
individual play, and a detailed note on the text used.
\ldots{} With more than 1,800 titles, Penguin Classics represents
a global bookshelf of the best works throughout history and
across genres and disciplines.
\\

\midrule

\textbf{Ground Truth}
&
Fiction; Science Fiction; Mystery \& Suspense;
Suspense \& Thriller; Cyber Punk
&
Classics; Fiction; Fiction Classics
\\
\midrule

\textbf{Only BCE}
&
Fiction; Science Fiction
&
Classics
\\
\midrule

\textbf{BCE + PathSimNCE}
&
Fiction; Science Fiction; Mystery \& Suspense;
Suspense \& Thriller
&
Classics; Fiction; Fiction Classics; Literary Collections
\\
\midrule

\textbf{BCE + DESC}
&
Fiction; Science Fiction; Mystery \& Suspense; Cyber Punk
&
Classics; Fiction; Fiction Classics; Literary Criticism
\\
\midrule

\textbf{Full Model}
&
Fiction; Science Fiction; Mystery \& Suspense;
Suspense \& Thriller; Cyber Punk
&
Classics; Fiction; Fiction Classics
\\

\bottomrule
\end{tabularx}
\caption{Predicted label sets of four model variants on two BGC test cases.}
\label{tab:case_analysis}
\end{table*}

\begin{figure*}[t]
\centering

\setlength{\fboxsep}{6pt}
\setlength{\fboxrule}{0.5pt}

\fbox{%
\begin{minipage}{0.95\textwidth}
\small
\raggedright

\textbf{Abstract:}

\noindent
``Title: Forever's Team. Text: The coaches and players of Duke University's
1977--78 team, who ended an otherwise successful season with their NCAA
championship loss to Kentucky, are traced from their time on the squad to
their present occupations. From the Hardcover edition.''

\vspace{5pt}

\textbf{Ground Truth Label:}

\noindent
["Nonfiction", "History", "Sports"]

\vspace{5pt}

\textbf{Hierarchical Label Paths:}

\noindent
[["Nonfiction", "History"], ["Nonfiction", "Sports"]]

\vspace{5pt}

\textbf{No Description:}

\noindent
``No Description'' removes the description alignment objective while keeping
PathSimNCE.

\vspace{5pt}

\textbf{Template Description:}

\noindent
["This document belongs to the category: Nonfiction, History.",

\noindent
"This document belongs to the category: Nonfiction, Sports."]

\vspace{5pt}

\textbf{LLM-generated Description:}

\noindent
["The hierarchical label path for Forever's Team consists of two levels:
Nonfiction and History. At the first level, Nonfiction encompasses works
that present factual information as opposed to fiction, which involves
invented or imaginative content. [...] The relationship between this level
and the content of Forever's Team is that the book provides a factual
recounting of events and individuals associated with Duke University's
1977--78 basketball team, emphasizing real-life outcomes and evolutions of
their careers.

\medskip

The second level, History, typically pertains to works that examine past
events, contexts, and developments. [...] Forever's Team aligns with this
label by tracing the historical trajectory of the 1977--78 Duke basketball
team, highlighting their journey from a significant NCAA championship loss
to their subsequent career paths. [...]",

\medskip

"The hierarchical label path for the given content consists of two levels:
Nonfiction and Sports. The first level, Nonfiction, [...] The abstract of
Forever's Team fits within this category as it describes the real-life
experiences and career trajectories of the coaches and players from Duke
University's 1977--78 basketball team. [...]

\medskip

The second level, Sports, is a more specific classification under the broader
category of nonfiction. This label pertains to works that focus on athletic
activities, events, teams, or personalities. [...] The content of Forever's
Team aligns with this label by concentrating on the sports domain,
specifically detailing the narrative of a collegiate basketball team. [...]

\medskip

In summary, the abstract's content is connected to the hierarchical labels
by providing a factual and historical account of a sports team, fitting
within the Nonfiction and Sports categories."]

\end{minipage}%
}

\caption{An example of the input-output format used to construct semantic supervision.}
\label{fig:semantic_example}

\end{figure*}

\begin{figure*}[t]
    \centering
    \includegraphics[width=\linewidth,keepaspectratio]{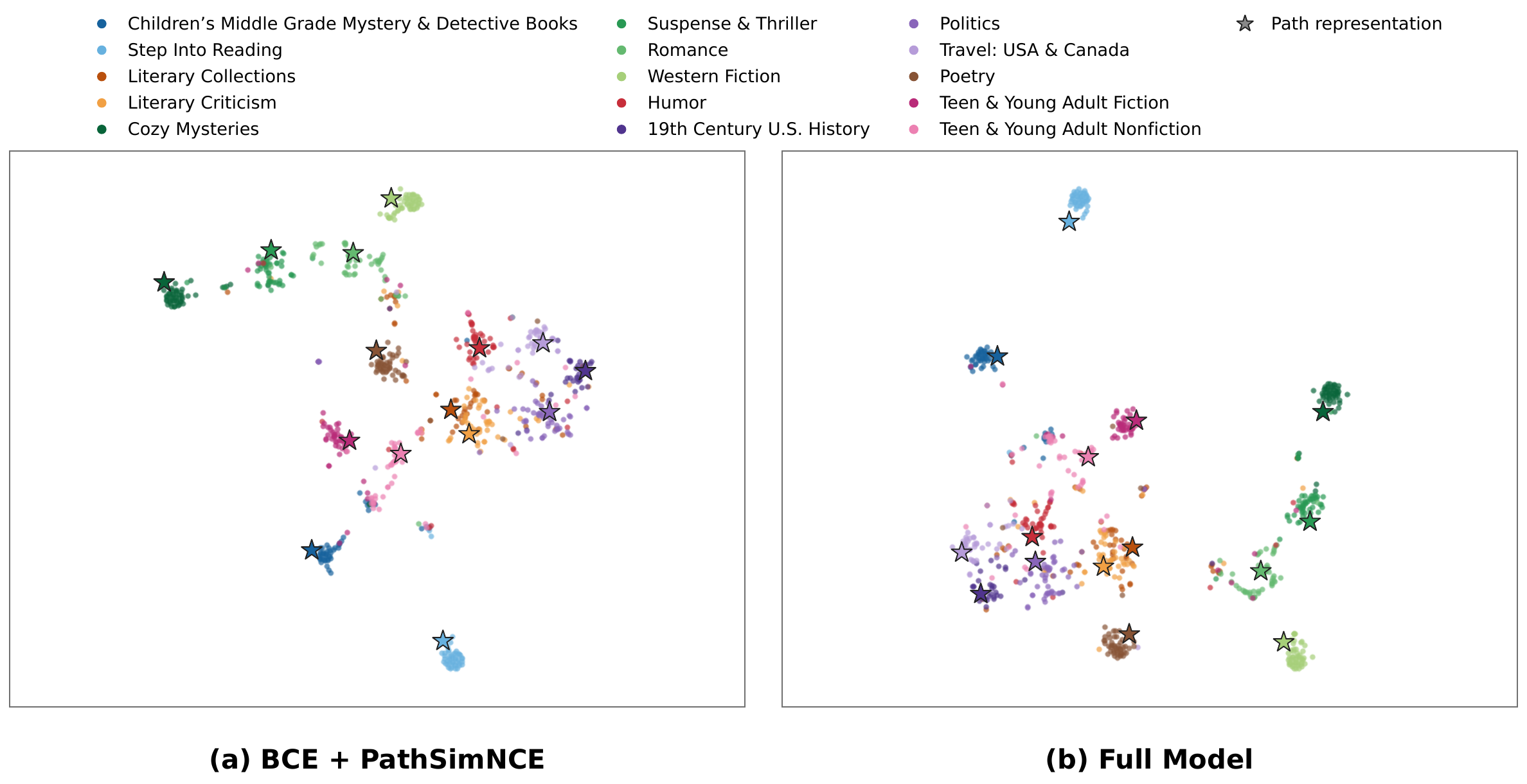}
    \caption{T-SNE visualization of text and path representations on the BGC dataset using the same 15 representative paths as in Figure~\ref{fig:4}. Circles of the same color denote text representations whose ground-truth label sets contain the corresponding path, while the star marker of the same color denotes that path representation. (a) BCE + PathSimNCE. (b) Full Model. The spatial proximity between each text group and its corresponding path representation provides a complementary view of the document--path alignment shown by the cosine-similarity heatmaps.}
    \label{fig:tsne}
\end{figure*}

\section{GPT Prompt}
\label{sec:supp_prompt}
We generate one semantic description for each gold text-path pair using GPT-4o and reuse these descriptions as auxiliary supervision during training. The procedure has three steps: (1) serialize the gold root-to-label path into a textual path string; (2) pair the path string with the source abstract; and (3) prompt the LLM to generate a concise explanation of how the document content matches the hierarchical path. We used the following prompt to generate the path descriptions using GPT-4o.

\begin{lstlisting}
Abstract:
"{abstract}"

Hierarchical Label Path:
{label_path_str}

Task:
Generate a detailed description that:
- Defines each label in the hierarchy and their relationships.
- Explains how the abstract content corresponds to each label.
- Avoids introducing external or unrelated information.
- Uses professional and academic language.
- Is approximately 200 words in length.
\end{lstlisting}

\section{Case Analysis}
\label{sec:case_analysis}

To qualitatively examine how the two auxiliary objectives affect the final predictions, Table~\ref{tab:case_analysis} presents two representative cases from the BGC test set. We compare the BCE baseline, the variants with PathSimNCE or description alignment, and the Full Model. All variants use the same standard multi-label prediction architecture and differ only in the supervision introduced during training.

In Case 1, the BCE baseline identifies only the coarse labels \textit{Fiction} and \textit{Science Fiction}. Adding PathSimNCE further recovers \textit{Mystery \& Suspense} and \textit{Suspense \& Thriller}, while description alignment additionally identifies \textit{Mystery \& Suspense} and the semantically specific label \textit{Cyber Punk}. The Full Model recovers the complete ground-truth label set. In Case 2, both auxiliary objectives recover the missing labels \textit{Fiction} and \textit{Fiction Classics}, but each single-objective variant also introduces one additional category. The Full Model retains the complete ground-truth label set without these extra predictions.

Overall, these cases illustrate that PathSimNCE and description alignment provide complementary structural and semantic supervision. Their joint optimization can lead to more complete and precise label predictions while retaining the standard multi-label classification architecture.

\section{Description Example}
\label{sec:supp_prompt_example}

The following example illustrates the input-output format used to construct semantic supervision (Figure~\ref{fig:semantic_example}).

\section{Performance across Label Frequencies}
\label{sec:label_frequency}

We further examine whether the proposed objectives benefit labels with different training frequencies. Labels are ranked by their numbers of training instances and divided into five equally sized percentile groups, ordered from the most frequent to the least frequent. Figure~\ref{fig:longtail} reports the Macro-F1 gain of each variant over the BCE baseline within each group.

All variants generally improve over BCE across the frequency groups, although the magnitude of the improvement varies. PathSimNCE provides consistent gains across both frequent and less frequent labels, while the effect of description alignment alone is comparatively modest. Notably, the full model achieves its largest improvement in the lowest-frequency group, suggesting that structural and semantic supervision are complementary for labels with limited training
examples.

\section{T-SNE Visualization}
\label{sec:tsne_visualization}




To complement the cosine-similarity heatmaps in Figure~\ref{fig:4}, we use t-SNE to visualize the learned text and path representations on BGC (Figure~\ref{fig:tsne}), using the same 15 representative paths. Same-colored circles denote text representations whose ground-truth label sets contain a given path, and the matching star denotes its path representation. Thus, both figures examine the same text groups and paths from different perspectives.

In Figure~\ref{fig:tsne}, text representations associated with each ground-truth path generally lie close to its path representation under both BCE+PathSimNCE and the Full Model. This spatial proximity complements the diagonal dominance in Figure~\ref{fig:4}'s heatmaps, where diagonal entries exceed off-diagonal similarity scores, jointly illustrating the learned document--path alignment. The Full Model preserves this alignment after incorporating description-based semantic supervision, consistent with the heatmap results.

\end{document}